%% file: main.tex
\pdfoutput=1
\documentclass[11pt]{article}
\usepackage[a4paper,margin=2.5cm]{geometry}
\usepackage[T1]{fontenc}
\usepackage{lmodern}
\usepackage{amsmath,amssymb}
\usepackage{graphicx}
\usepackage{booktabs}
\usepackage{tabularx}
\usepackage{caption}
\usepackage[hidelinks]{hyperref}
\usepackage{enumitem}
\usepackage{microtype}
\setlist{nosep,leftmargin=*}
\newcommand{\astra}{GPT-6 Astra}
\hypersetup{pdftitle={Computation, not weight access: large language models as auditors of clinical prediction networks},pdfauthor={Zhiyu Liu, Xinyu Yang}}

\title{Computation, not weight access: large language models as auditors of clinical prediction networks}
\author{Zhiyu Liu\textsuperscript{1} \qquad Xinyu Yang\textsuperscript{2,*}\\[6pt]
{\small \textsuperscript{1}Tianjin Medical University, Tianjin, China; \href{mailto:07eb1f@gmail.com}{07eb1f@gmail.com}}\\
{\small \textsuperscript{2}Tianjin Medical University General Hospital, Tianjin, China; \href{mailto:yangxinyu@tmu.edu.cn}{yangxinyu@tmu.edu.cn}}\\
{\small \textsuperscript{*}Corresponding author}}
\date{November 2026}

\begin{document}
\maketitle

\begin{abstract}
\noindent Frontier large language models (LLMs) have been reported to read, edit and write the weights of small neural networks, raising the prospect of auditing clinical prediction models from their files alone. Public demonstrations, however, mixed reading raw weights with running code. We separated these conditions in internally pre-specified simulation experiments. Tanh networks of 105 to 38,401 weights were trained on simulated cohorts of patients with subarachnoid haemorrhage, each with a planted relationship or none, and one network was trained on labels that under-recorded deaths in a subgroup. \astra{} and four other GPT models answered single prompts under a default or an explicit system instruction, either with the weights in the prompt and no tools, or with a calculation-only tool that exposed the weights (white box) or only the network's predictions (black box). Without tools, Astra recovered planted relationships in 353-weight networks but became unreliable at 3,457 weights, and under the default instruction it declined 17 of 21 audit and repair tasks as unidentifiable; the explicit instruction removed these refusals. With the tool, it recovered the planted relationship in all 39 runs on networks of up to 38,401 weights, raised no false alarm in 21 runs on null networks, and found the biased subgroup in 6 of 6 runs. Black-box access performed as well as white-box access, and other GPT-6 models read as accurately. A well-designed script matched the comparator models on the audit and the best LLM repair. For auditing these networks, what mattered was computation and elicitation, not access to the weights.
\end{abstract}

\section{Introduction}

In November 2026, Takehi and colleagues reported that \astra{} could tell what small networks had been trained to do by reading their raw weights, write working policy networks from a description of the task, and repair a 0.8-billion-parameter language model by editing a small fraction of its weights, all without gradient-based training \cite{takehi2026}. Independent re-runs reproduced the identification of game-playing networks and showed that it fell to chance once the weights were shuffled within their matrices \cite{turing2026}. The report also drew critiques. The most pointed one held that the headline combined two different conditions under one claim: reading raw weights of a small network without any tool, and analysing a larger model by writing and running code \cite{honnibal2026}. Others argued that the small-network results could be matched by simple inspection \cite{goedecke2026}, that the reported gain in mathematics rested on a weak baseline setting \cite{toledo2026}, and that reading raw weights helped little in predicting what a weight edit would do \cite{singh2026}. Whether the ability lies in reading weights or in computing with them has practical consequences, because the two require different access to a model.

Clinical prediction models make the question concrete. Hospitals, regulators and reviewers increasingly receive models without the data they were trained on, and such models can encode problems that headline performance does not reveal. One example is a label that records outcomes unevenly across patient groups \cite{obermeyer2019}. Post hoc explanation of black-box clinical models has well-known limits \cite{rudin2019,ghassemi2021}. An auditor that could describe what a model computes, find where it departs from what is plausible, and correct it would be useful. Nusrat and Nusrat recently showed that Claude Opus 5.5, reading only the weights of small rule-based radiotherapy-planning networks, named a planted flaw in 16 of 18 networks, although simple programs were faster and judging a healthy network was harder \cite{nusrat2026}.

We asked which conditions such performance depends on. We trained clinical prediction networks on simulated data with known ground truth and gave them to \astra{} and four other GPT models under conditions that the public debate had mixed: the weights in the prompt with no tools, a calculation-only tool holding the weights, and the same tool exposing only the network's predictions. We also varied the system instruction, because many first replies declined tasks rather than attempting them. Our contributions are:
\begin{itemize}
\item a factorial evaluation of instruction and access on clinical-style networks of 105 to 38,401 weights, with five models, pre-specified scoring rules, negative controls and repeats;
\item evidence that reading weights without tools works only for small networks and varies between repeats, whereas computation makes reading and auditing accurate and stable at every size tested, with or without access to the weights;
\item evidence that the default single-reply set-up produced refusals framed as non-identifiability, and that one paragraph of system instruction removed them; without tools, false alarms in our controls arose only when the prompt stated a reference, under either instruction;
\item a comparison with scripted analyses and edits that use no LLM, which places the LLM's added value in designing and carrying out the audit rather than in capabilities beyond a well-specified script;
\item the release of all networks, prompts, replies, logs and code.
\end{itemize}

\section{Related work}

\paragraph{Weight-space programming.} The original report \cite{takehi2026} demonstrated reading, writing and editing of network weights by \astra{}; small networks were read from raw weights alone, while the edits of larger language models used code and repeated evaluation. Re-runs reproduced the reading of game-playing networks with \astra{} and Claude Opus 5.5 and showed its dependence on intact weights \cite{turing2026}. Critics questioned the framing of these results as native weight reading \cite{honnibal2026,goedecke2026}, the baseline behind the mathematics result \cite{toledo2026}, and how much reading raw weights adds to predicting the effect of an edit \cite{singh2026}.

\paragraph{Language models that read weights.} Closest in aim, Weight Oracles fine-tune a 3-billion-parameter language model to diagnose properties of small transformers, such as backdoors, from their raw weights, with parameter-free operations delegated to deterministic code \cite{kabra2026}; whether off-the-shelf frontier models could do the same was left open. Diff Interpretation Tuning trains models to describe their own fine-tuning changes in natural language \cite{goel2025}. We study off-the-shelf frontier models without fine-tuning, give them a calculator that they call themselves, and separate access to the weights from the ability to compute.

\paragraph{Automated interpretability agents.} Language models have been used as agents that explain other systems by experimenting on them. FIND provides synthetic functions with known descriptions and found that an LLM agent with black-box access can infer much of a function's structure, although its descriptions captured global behaviour better than local details \cite{schwettmann2023}. MAIA equips a multimodal model with tools to run experiments on vision models and describe their components \cite{shaham2024}. Our black-box condition is an agent of this kind; our question is whether access to the weights adds anything once such experiments are possible.

\paragraph{Auditing clinical prediction models.} Clinical algorithms can inherit biases from how their labels were recorded \cite{obermeyer2019}, and explanations of black-box models can be unfaithful or misleading \cite{rudin2019,ghassemi2021}. LLMs can read glass-box additive models and flag effects that contradict prior knowledge \cite{lengerich2023,bordt2024}, without testing whether those effects are real. The closest study to ours, concurrent and independent, audited small rule-based clinical decision networks from their weights alone, without tools \cite{nusrat2026}. We add networks trained on noisy probabilistic outcomes, sizes an order of magnitude larger, conditions with and without computation and weight access, and several models.

\section{Methods}

\subsection{Overview}

Table~\ref{tab:design} summarises the design. All networks were trained on simulated data, so the true relationships were known. Each run was one prompt and one reply, and every reply was scored by a script against the ground truth with rules written before the runs.

\begin{table}[t]
\centering
\caption{Study design at a glance.}
\label{tab:design}
\small
\begin{tabularx}{\textwidth}{@{}lX@{}}
\toprule
Element & Levels \\
\midrule
Networks & Tanh multilayer perceptrons trained on simulated cohorts: 105 weights (11 inputs); S, 353; M, 1,217; L, 3,457; D3 (three hidden layers), 2,817; XL, 11,009; XXL, 38,401 (20 inputs); and a 209-weight network trained on biased labels, with an unbiased counterpart \\
Planted relationships & Threshold; interaction; each at half strength; none (null). In the 105-weight networks: a threshold, a hypertension-by-occupation interaction, an age dip \\
Tasks & Read (explain and write a surrogate model); audit (find subgroups whose predicted risk departs from what their other characteristics imply); repair (edit the weights) \\
Instruction & Default (one sentence); explicit (open choices left to the model; Appendix~\ref{app:instruction}) \\
Access & Weights in the prompt, no tools; calculator with the weights (white box); calculator with predictions only (black box) \\
Models & \astra{} (main); \texttt{gpt-6.1-sol}, \texttt{gpt-6-sol}, \texttt{gpt-6-luna}, \texttt{gpt-5.6-sol} \\
Baselines without an LLM & Exact scan of each network; scripted audits (regression tree; exhaustive search); scripted repairs (gradient fine-tuning; a scripted gate) \\
Repeats & Three runs per cell of the reading, audit and repair tables; one per cell for the other models \\
\bottomrule
\end{tabularx}
\end{table}

\subsection{Simulated networks}

Each simulated cohort had 20,000 adults with aneurysmal subarachnoid haemorrhage. The ten clinical inputs were coded as in a typical admission dataset: age (in decades from 60), sex, poor clinical grade (WFNS 4--5), thick haemorrhage (Fisher 3--4), intraventricular and intracerebral haemorrhage, hypertension, aneurysm treatment during admission, urban residence and farming occupation. Their distributions and the coefficients of the outcome model were set by hand to clinically plausible values and were not estimated from any dataset. Ten neutral laboratory values ($l_1$ to $l_{10}$, uniform on 0--10) were added, so that medical knowledge could not help with relationships planted on them. The outcome, death within one month, was drawn from a logistic model with a common clinical part and at most one planted relationship: a threshold, $+0.9\max(l_3-6,0)$, or an interaction, $+1.6\,\sigma(l_5)\sigma(l_8)$ with $\sigma(x)=1/(1+e^{-2(x-5)})$, so that risk rose only when both values were high; each also at half strength; or none.

Networks were multilayer perceptrons with tanh units, trained with scikit-learn (Table~\ref{tab:design}). Before use, each network had to pass a stimulus check: it had to carry at least half of the planted strength, and a null network could carry no signal of 0.3 or more at the planted locations. Training used Adam with early stopping and fell back to L-BFGS with an L2 penalty when a network failed the check. A further set of three bootstrap networks per condition, trained on resamples of one cohort, was given to the model together, with the request to report only the structure they shared. An earlier set of 105-weight networks (11 inputs, eight hidden units) carried a threshold on one laboratory value, an interaction between hypertension and farming, or a dip in risk between ages 50 and 60.

For auditing, a 209-weight network (11 inputs, 16 hidden units) was trained on labels in which the recorded risk of women farmers over 65 (about 4\% of patients) was lowered by 3.0 on the log-odds scale, mimicking deaths that go unrecorded when patients are taken home to die. Its log-odds in that subgroup were 1.60 lower than those of a network trained on unbiased labels, which served both as the reference for scoring and as a negative control.

\subsection{Tasks and prompts}

Each prompt named the outcome and the population (adults admitted with aneurysmal subarachnoid haemorrhage), gave the inputs and their coding, the architecture and the weights (as JSON with four significant figures), and asked for a brief explanation ending in a JSON block (Appendix~\ref{app:prompt}).
\begin{itemize}
\item \textbf{Read.} The model explained what the network computes and wrote a surrogate model of linear, hinge and product terms, together with a list of relationships that were ``not standard medical knowledge''.
\item \textbf{Audit.} The model listed subgroups, defined by conditions on the inputs, whose predicted risk was systematically lower or higher than their other characteristics implied, with the size of the distortion. In one variant the prompt stated the reference: the network's additive part over a described population.
\item \textbf{Repair.} The model returned the complete edited network in the same shape. In one variant it had to find the distorted subgroup itself; in others the subgroup was given, with the target either to remove its departure from the additive part or to raise its log-odds by exactly 1.6 while leaving everyone else unchanged. Two variants stated a tolerance.
\end{itemize}
A reply without a usable JSON block, or one that declared the requested fields empty, was recorded as declined.

\subsection{Instruction and access}

\textbf{Instruction.} The default system instruction was one sentence: ``Answer the user's request.'' Because many replies under it declined tasks that required a choice the prompt had left open, we added an explicit instruction while keeping every prompt unchanged (Appendix~\ref{app:instruction}). It told the model that it had one reply and no one to ask; to choose the most defensible option for anything left open (a reference, a population, a tolerance or a size), name it and continue; to give the closest approximation, and say how close it was, when a request could not be met exactly; and not to add disclaimers. This follows published guidance for GPT-6 models, which notes their tendency to ask when more input would change the result \cite{openai2026}.

\textbf{Access.} Without tools, the weights were in the prompt and every tool was disabled. With the calculation-only tool (hereafter, the calculator), the model had a single tool, \texttt{run\_python}, served by a local Model Context Protocol server, which ran NumPy code with the network preloaded: with its weight arrays and a prediction function (white box), or with the prediction function only (black box). The calculator refused imports, names beginning with an underscore and NumPy's file functions, ran each call in a fresh process in an empty folder with a 60-second limit, and accepted only simulated networks. In the reading prompts with the calculator, the weights were held by the tool rather than printed. Every tool call was logged and audited.

\subsection{Models and harness}

The main model was \astra{} (\texttt{gpt-6-astra}) at its maximal reasoning effort. The comparators were \texttt{gpt-6.1-sol}, \texttt{gpt-6-sol}, \texttt{gpt-6-luna} and \texttt{gpt-5.6-sol}, at the same effort. All ran through the Codex command-line interface (version 0.166.0) in non-interactive mode, one prompt and one reply per run, each in a new empty folder with the user's configuration ignored, a read-only sandbox and web search off. Comparator models received the same prompts, byte for byte.

\subsection{Outcomes}

\textbf{Reading.} The strength of a planted relationship was computed for the network and for the surrogate on the same 5,000 test inputs: for the threshold, the effect of $l_3$ from 5 to 9 minus its effect from 1 to 5; for the interaction, the interaction contrast of $l_5$ and $l_8$ at 2 and 8. A relationship counted as recovered if the surrogate's strength had the network's sign and at least half its size. On a null network, a surrogate strength of 0.5 or more at either planted location was a false alarm. We report the strength ratio (surrogate over network) and the $R^2$ of the surrogate against the network's log-odds. As a further check, we counted surrogate effects of 0.5 or more at 65 locations that were zero in the generating model: the range effect and curvature of each laboratory value, and each pairwise interaction between them. For the 105-weight networks we applied the rules written for that experiment, which judge the shape of the surrogate.

\textbf{Auditing.} A reported subgroup counted as found if its Jaccard overlap with the true subgroup on the test inputs was at least 0.5 and its stated shift had the right sign and at least half the network's actual shift. On the unbiased network, any subgroup with a stated shift of 0.5 or more was a false alarm.

\textbf{Repair.} We report the mean change in log-odds inside the subgroup, the mean absolute change elsewhere, and the remaining difference from the unbiased network inside the subgroup.

\subsection{Baselines without a language model}

For reading, an exact scan of each network at the 65 locations shows what a script reading the network perfectly would report. For auditing, a script with prediction access only fitted an additive model to the network's log-odds over the balanced synthetic population that Astra had chosen in its calculator runs (ages 18 to 90, binary inputs equally likely), then searched the residuals in two pre-specified ways: a regression tree (depth 3, leaves of at least 1\%) and an exhaustive search over conjunctions of up to three binary conditions with an optional age cut. For repair, we compared gradient fine-tuning of all parameters with a scripted gate, in which the hidden unit whose removal cost least became a steep gate for the subgroup and the output layer was refitted by least squares.

\subsection{Pre-specification and statistics}

The plan, stimuli and scoring code of every experiment were committed to a version-controlled repository before its first run; an addendum scoring claims at the 65 zero locations was committed before any affected reply was read (Appendix~\ref{app:prespec}). This was internal pre-specification, not a public registration, and the design was adaptive: later conditions were designed after earlier results were known. Results are descriptive; proportions are given with Wilson 95\% intervals \cite{wilson1927}.

\section{Results}

\subsection{Without tools, reading works for small networks only}

\begin{figure}[t]
\centering
\includegraphics[width=\textwidth]{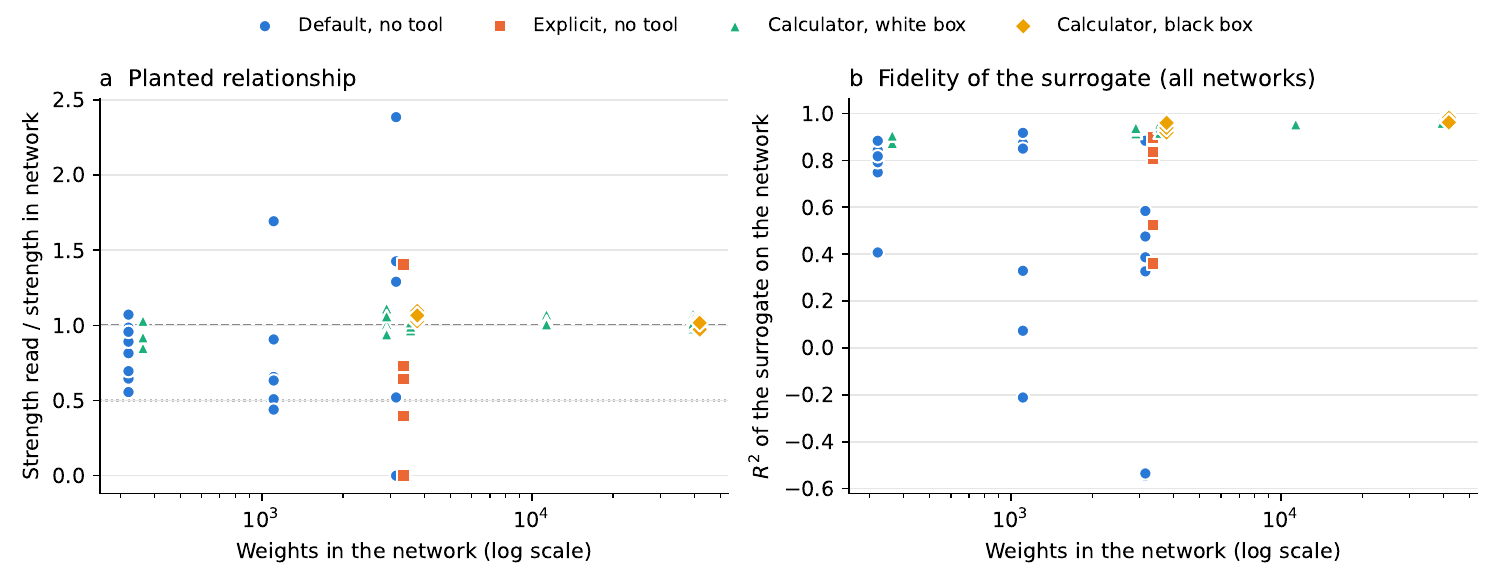}
\caption{Reading planted relationships across network sizes (\astra{}). (a) Strength of the planted relationship in the model's surrogate divided by its strength in the network; the dashed line marks equality and the dotted line the pre-specified threshold of one half. (b) $R^2$ of the surrogate against the network's log-odds, including null networks. Default and explicit refer to the system instruction; the calculator held either the weights (white box) or only a prediction function (black box). Points are single runs, offset slightly on the horizontal axis.}
\label{fig:reading}
\end{figure}

Under the default instruction and without tools, \astra{} recovered the planted relationship in all 12 runs on the 353-weight networks, including the half-strength versions, with surrogate $R^2$ of 0.41--0.89 and no false alarm in three runs on the null network (Table~\ref{tab:reading}, Figure~\ref{fig:reading}). At 1,217 weights it recovered the interaction in all three runs but the threshold in two of three (strength ratios 0.44 to 1.69), and its null surrogates had $R^2$ of $-$0.21 to 0.87. At 3,457 weights it recovered the threshold in all three runs (strength ratios 0.52 to 1.43) but the interaction in only one of three, with strength ratios of 0.00 to 2.39 and surrogate $R^2$ of 0.33--0.39; the three null surrogates had $R^2$ of $-$0.54 to 0.88. Repeats of the same cell often differed widely, and single runs took 23 to 62 minutes.

The failing replies explained themselves. Astra wrote that it could not carry out the numerical sweep it had planned, linearised the network around a midpoint (age 60, laboratory values 5, binary inputs 0.5) and reported the local gradient. The gradient was correct, but a local linearisation is a poor global surrogate, and its linear extrapolation produced spurious effects at up to seven zero locations.

Astra read the function rather than the pattern of numbers. When the hidden units of two 105-weight networks were permuted and half of them sign-flipped, which changes every weight but not the function, it read the same relationships with the same strengths (2.44 against 2.35 before; 1.71 against 1.72), with $R^2$ of 0.97 and 0.96.

\begin{table}[t]
\centering
\caption{Reading planted relationships (\astra{}). Each cell gives the strength ratio and the surrogate $R^2$ (range over runs) and, in brackets, the number of runs in which the relationship was recovered by the pre-specified rule; for null networks, $R^2$ and the number of false alarms.}
\label{tab:reading}
\resizebox{\textwidth}{!}{\input{tables/reading.tex}}
\end{table}

\subsection{The default instruction produced refusals}

Under the default instruction, 17 of 21 audit and repair runs declined (95\% interval 0.60--0.92). Astra argued that a distortion ``implied by other characteristics'' is not identifiable without a reference, that a stated set of marginal distributions does not determine the joint distribution, or that a tanh network, being continuous in age, cannot change only patients above 65 exactly. These arguments are correct as mathematics, but they answer a different question from the one asked. Of the four runs that answered, three had an additive reference stated in the prompt: one found the subgroup and two raised false alarms on the unbiased network. The fourth, a repair, returned the network almost unchanged (Tables~\ref{tab:audit} and~\ref{tab:repair}).

Under the explicit instruction, with every prompt unchanged, none of 24 audit and repair runs declined (95\% interval 0.00--0.14). Its effect on reading without tools was smaller. On the 3,457-weight networks the planted relationship was recovered in four of six runs, as often as under the default instruction, though the interaction more often (two of three runs against one) and the threshold less often (two of three against three), with surrogate $R^2$ of 0.36--0.90. The null surrogates, however, became consistently faithful, with $R^2$ of 0.84--0.90 against $-$0.54 to 0.88 (Table~\ref{tab:reading}), and on three bootstrap null networks given together, in one run under each instruction, the surrogate $R^2$ rose from $-0.24$ to 0.88. On the unbiased audit network the largest departures it reported were 0.25, 0.30 and 0.40 in the three runs, all below the false-alarm threshold. Stating an additive reference in the prompt did not help. Every run that answered with the reference stated raised false alarms on the unbiased network, two of three under the default instruction (the third declined) and all three under the explicit instruction, each time in patients with a very low value of the laboratory input, the edge region where the network is genuinely non-additive.

The explicit instruction also had costs. In all three runs asked to raise one subgroup's log-odds by exactly 1.6, Astra chose to minimise the worst-case error over patients; because a continuous network cannot reproduce an exact step, adding 0.8 to every patient is optimal under that criterion, and that is what it returned. With a tolerance also stated, it changed the network from 16 to 17 hidden units in two of three runs, against the instruction to keep the shapes, and said so.

\subsection{With the calculator, size was no obstacle and the weights were not needed}

With the calculator, \astra{} recovered the planted relationship in all 39 runs that had one (95\% interval 0.91--1.00), on networks of 353 to 38,401 weights, with strength ratios of 0.85--1.11 and surrogate $R^2$ of 0.88--0.98 (Table~\ref{tab:reading}, Figure~\ref{fig:reading}). It raised no false alarm in 21 runs on null networks (0.00--0.15) and made at most one spurious claim per run at the 65 zero locations. Repeats agreed closely; for example, the strength ratio for the 3,457-weight threshold network was 1.02--1.07 across three white-box and three black-box runs. Runs took a median of 13 minutes (range 9 to 30) with 7 to 15 tool calls.

Access to the weights made no measurable difference. In the 18 black-box runs, in which the calculator exposed only the network's predictions, strength ratios were 0.97--1.10 and $R^2$ 0.92--0.98, matching the white-box runs cell by cell.

\subsection{Other models}

Reading small networks was not particular to Astra (Table~\ref{tab:models}). On the 105-weight networks without tools, \texttt{gpt-6.1-sol} recovered 3 of 3 planted relationships with $R^2$ of 0.96--0.97, as Astra did (6 of 6; 0.94--0.97). \texttt{gpt-5.6-sol}, of the previous generation, recovered 3 of 3 by the shape rules but with poorer fits, and \texttt{gpt-6-sol} and \texttt{gpt-6-luna} recovered 2 of 3. With the calculator, all three GPT-6 comparators read the 3,457- and 38,401-weight networks as accurately as Astra (interaction strength ratios 0.89--1.13; null $R^2$ 0.92--0.98), although \texttt{gpt-6-luna} needed about twice as many tool calls.

\subsection{Auditing a network trained on biased labels}

\begin{figure}[t]
\centering
\includegraphics[width=0.92\textwidth]{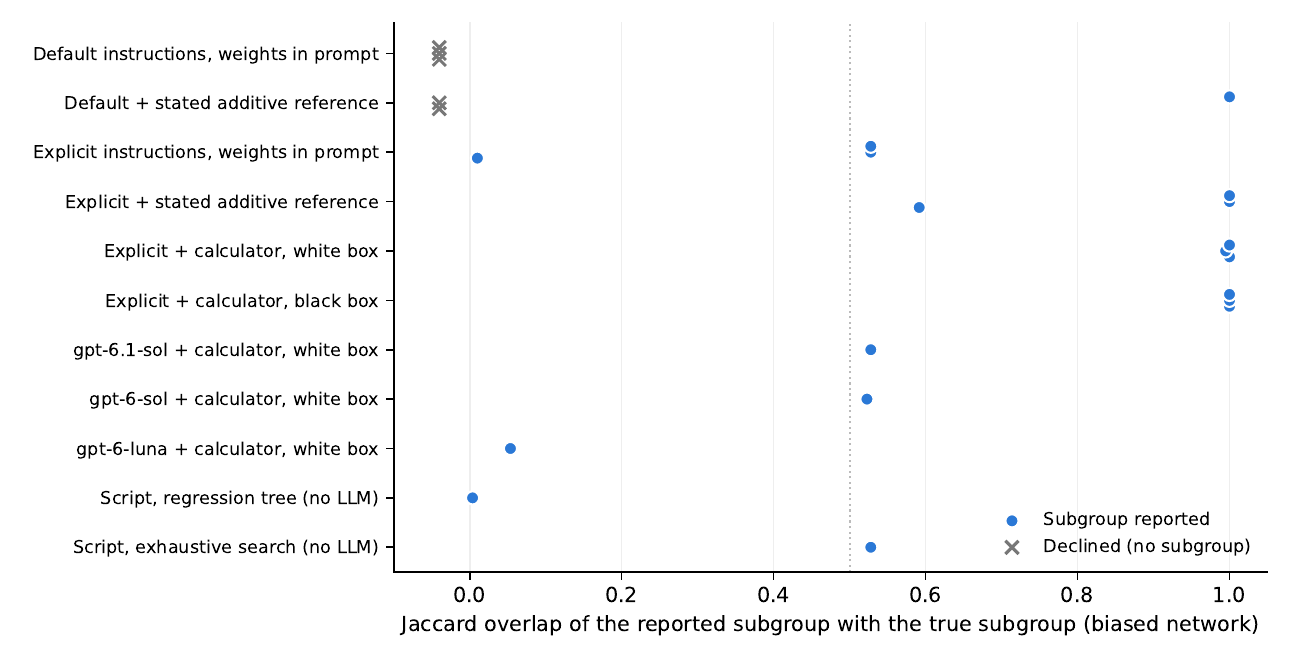}
\caption{Recovery of the biased subgroup. Each point is one run on the biased network: the Jaccard overlap between the true subgroup (women farmers over 65) and the reported subgroup closest to it. Crosses mark runs that declined to report a subgroup; the dotted line marks the pre-specified threshold of 0.5. Rows without a model name are \astra{}.}
\label{fig:audit}
\end{figure}

With the calculator, \astra{} found the biased subgroup in all six runs, three white-box and three black-box (95\% interval 0.61--1.00). Its subgroups coincided with the true one (Jaccard 0.995--1.000), and its stated shifts ranged from $-$1.60 to $-$2.00 against the network's $-$1.60 (Table~\ref{tab:audit}, Figure~\ref{fig:audit}). It chose its own reference without being told: an additive model fitted over a balanced synthetic population. On the unbiased network it raised one false alarm in six runs: a white-box run flagged patients with a very low value of the laboratory input (shift $+$0.53, just above the threshold), the edge region where the network is genuinely non-additive. The three black-box runs reported nothing.

The comparator models were less precise. \texttt{gpt-6.1-sol} and \texttt{gpt-6-sol} found the subgroup but placed the age boundary at 70 (Jaccard 0.53 and 0.52); \texttt{gpt-6-sol} and \texttt{gpt-6-luna} each raised a false alarm, and \texttt{gpt-6-luna} missed the subgroup. Without tools, under the explicit instruction, Astra found the subgroup in two of three runs, both times with the age boundary at 70 (Jaccard 0.53); in the third it named the right groups (``older farmers, especially women'') but defined the subgroup as a narrow profile that almost no patient met.

A script did as well as the comparator models. With prediction access only, the exhaustive search found the subgroup in under a second, with the same shifted boundary (women farmers over 70; Jaccard 0.53; shift $-$1.35); the regression tree did not, settling on the edge region of a laboratory value. Neither raised a false alarm. Two caveats apply: we designed the search space knowing the form of the planted subgroup, whereas Astra designed its own search and placed the boundary exactly.

\begin{table}[t]
\centering
\caption{Auditing the biased network and the unbiased control. Found: pre-specified rule (Jaccard $\geq$ 0.5, correct sign, at least half the network's shift). False alarm: a stated shift of 0.5 or more on the unbiased network. Rows without a model name are \astra{}.}
\label{tab:audit}
\resizebox{\textwidth}{!}{\input{tables/audit.tex}}
\end{table}

\subsection{Repair succeeded only partly}

Editing the weights, the one task that requires them, was the weakest (Table~\ref{tab:repair}). Under the default instruction eight of nine repair runs declined, and the ninth returned the network almost unchanged. Under the explicit instruction without tools, Astra chose a wrong subgroup in all three runs when it had to find the distortion itself (Jaccard 0.01). With the subgroup given and a tolerance stated, it turned one hidden unit into a gate for the subgroup and compensated for that unit's former contribution through the others, in one run after merging two units to free it; the remaining bias in the subgroup fell from $-$1.61 to between $-$0.37 and $-$0.20, and other patients' log-odds changed by only 0.07 to 0.09 on average. Asked for an exact $+$1.6, it gamed the criterion when no tolerance was stated and broke the shape constraint in two of three runs when one was, as described above; in the third it kept the shapes but missed the tolerance, with a worst-case error of about 0.83 that it reported itself. With the calculator, in all three runs it refitted 15 units to the original function and used the sixteenth as a gate: the subgroup's log-odds rose by 1.48 to 1.49, but other patients' changed by 0.20 to 0.21 on average, a shortfall against the requested tolerance that it reported itself.

The scripted gate, computed in 0.2 seconds and changing 29 of 209 parameters, did slightly better (1.51 in the subgroup, 0.13 elsewhere). Gradient fine-tuning changed 207 of 209 parameters and undershot (1.20).

\begin{table}[t]
\centering
\caption{Repairs. Change in subgroup: mean change in log-odds for women farmers over 65 (target $+$1.6, or removal of the departure). Bias after: remaining difference from the unbiased network inside the subgroup (before: $-$1.60). Invalid: an answer that broke the shape constraint. Rows above the line are \astra{}.}
\label{tab:repair}
\resizebox{\textwidth}{!}{\input{tables/repair.tex}}
\end{table}

\begin{table}[t]
\centering
\caption{Other models. Left: the 105-weight networks without tools (pre-specified shape rules; three planted relationships). Right: the calculator with white-box access on the 3,457- and 38,401-weight networks and the audit.}
\label{tab:models}
\resizebox{\textwidth}{!}{\input{tables/models.tex}}
\end{table}

\subsection{Faithful to the network, noise included}

Astra described the networks it was given, including their noise. With the calculator, its lists of ``unexpected'' relationships in null networks contained three to eight entries per run. These were real, small wiggles of the networks themselves, changes of about 0.3 to 0.5 across a laboratory value's whole range, and they were correctly sized; the surrogates did not exaggerate them. A reader who did not look at the sizes could still mistake them for findings. Some artefacts were systematic: networks trained on data with a threshold on $l_3$ carried negative interactions between $l_3$ and other laboratory values that were absent from the generating model, and three networks trained on bootstrap resamples of one cohort agreed on them (for example $-$0.55, $-$0.57 and $-$0.85 for $l_3\times l_{10}$). Agreement across resamples therefore cannot separate structure in the data from artefacts of training.

\section{Discussion}

\subsection{Principal findings}

Three findings stand out. First, reading a network's weights without tools worked, but only for small networks of the order of $10^3$ weights, and results varied between repeats; beyond that size the model said, correctly, that it could not do the computation and fell back on a local approximation. Second, with a calculation-only tool, reading and auditing were accurate and stable at every size we tried, up to 38,401 weights, and access to the weights added nothing measurable: given only the network's predictions, Astra read and audited as well as with the weights. Third, the default single-reply set-up made the model decline most audits and repairs, and one paragraph of system instruction, with no change to the prompts, removed the refusals; without tools, false alarms in our controls arose only when the prompt stated a reference, under either instruction.

\subsection{Relation to the debate on weight-space programming}

The critique that tool-free weight reading and code-assisted analysis had been conflated \cite{honnibal2026} is borne out: the two behaved very differently, and the second did not depend on the weights at all. The limited value of reading raw weights for predicting edit effects \cite{singh2026} fits our size ladder. Our findings agree with Nusrat and Nusrat \cite{nusrat2026} that frontier models can find planted problems in small clinical networks from the weights alone, that declaring a network healthy is harder, and that simple programs are often as good and faster. In their study, a black-box condition without tools and with a limited number of requests did worse than weight reading; in ours, a black-box condition with a calculator did as well. Together, the two studies suggest that performance is decided by the budget for computation and queries, not by whether the weights are visible. Our results also bear on the question left open by Weight Oracles \cite{kabra2026}: off-the-shelf frontier models with a calculator read networks an order of magnitude larger than its targets, although simpler in architecture, and did as well without the weights. This places the useful capability in the tradition of automated interpretability agents that explain systems by experimenting on them \cite{schwettmann2023,shaham2024}. Unlike the agents evaluated in FIND, which captured global behaviour better than local details, the calculator runs here recovered planted local structure such as the position of a threshold; the low dimension and smoothness of clinical tabular networks may make this easier.

\subsection{Implications for auditing clinical models}

An LLM with a calculator and query access carried out a routine audit end to end: it chose a reference, searched for departures, sized them and described them. Because it did not need the model file, such an audit would not require a vendor to disclose its weights. The comparison with scripts tempers the claim. Once a person knows what kind of problem to look for, a few lines of code do as well; what the LLM adds is that it designs and carries out the search itself and reports in plain language, and in our tests Astra was also the most precise. Editing the weights remained the weakest task: constraints that are trivial for a script, such as keeping the shape of the network or staying within a tolerance, were where the model gamed the criterion or broke the rule. Faithful reading is also not the same as a trustworthy finding, because the model reported the networks' own noise and training artefacts as faithfully as their planted structure.

\subsection{Implications for evaluating LLM capabilities}

Reported abilities depended on the instruction and the tools as much as on the model. A single-reply protocol removes the model's preferred way of resolving an open choice, which is to ask a question, and the model then argues that the task cannot be done. Without tools, the explicit instruction alone removed the refusals and made the surrogates of 3,457-weight null networks consistently faithful, although it did not raise the recovery of planted relationships. Evaluations should therefore report the instruction and tools used, count refusals separately from wrong answers, and check any change of elicitation against negative controls, as we did with the null and unbiased networks.

\subsection{Limitations}

All networks were small multilayer perceptrons trained on simulated data with hand-set parameters, and the planted relationships and the label bias were of simple, known forms; the results may not transfer to models trained on real clinical data, to other architectures such as gradient-boosted trees, or to problems of unknown form. The models came mainly from one family and were accessed through one harness; three runs per cell give wide intervals, and the comparisons with other models had one run per cell. The scripted audit's search space was designed with knowledge of the planted subgroup's form, which favours the script. The prompts described each network as a clinical prediction model, which may have influenced the language models' priors. The models are commercial services that may change, and no open-weight model was evaluated. Finally, the rules were pre-specified internally rather than publicly registered, and the design was adaptive (Appendix~\ref{app:prespec}).

\section{Conclusion}

Without tools, frontier LLMs read small clinical prediction networks from their weights but not larger ones. With a calculation-only tool they read and audited networks of up to 38,401 weights accurately and stably, and they did as well without seeing the weights. For auditing clinical prediction models, the useful capability is automated experimentation, which in our tests performed on a par with well-designed scripts. Any claim about what these models can do should state how the task was posed and which tools were available.

\section*{Data and code availability}
All simulated networks, prompts, replies, event logs and evaluation code are available at \url{https://github.com/Chth1z/llm-clinical-network-audit}. No patient data were used.

\section*{Ethics}
This study used only simulated data and involved no human participants.

\section*{Use of AI}
The GPT models listed in the Methods were the objects of study. Claude Opus 5.5 (Anthropic), used through Claude Code, assisted with the experimental code, the analyses and the drafting of this manuscript under the authors' direction. The authors checked the results and take full responsibility for the content.

\section*{Competing interests}
The authors declare no competing interests.

\section*{Funding}
This work received no specific funding.

\appendix

\section{Pre-specification record and deviations}
\label{app:prespec}

For every experiment, the plan with its decision rules, the stimuli (networks and prompts) and the scoring code were committed to a private version-controlled repository before the first run (Table~\ref{tab:prespec}); the decision rules are restated in the Methods.

\begin{table}[h]
\centering
\caption{Plan commits and the start of the first run of each experiment.}
\label{tab:prespec}
\resizebox{\textwidth}{!}{\input{tables/prespec.tex}}
\end{table}

\noindent Deviations from the plans:
\begin{enumerate}
\item Up to 14 runs were started at once, more than planned, which led to ``model at capacity'' errors in three runs. Seven batches that had run for two to ten minutes were stopped and re-queued, and the failed runs were repeated with identical prompts. Afterwards a queue limited the machine to six concurrent runs and retried runs automatically when the service was busy.
\item In the repeats that brought every cell to three runs (last row of Table~\ref{tab:prespec}), two runs failed because the connection to the service dropped in each of five attempts, before any reply. Their records were kept and the runs were repeated with identical prompts.
\item Replies that declared the requested fields empty, or contained no JSON block, were scored as declined, as the plans specified; the parser was changed to record them as such, without changing the rules.
\item The addendum scoring claims at the 65 zero locations, and the exact scan, were added after the size-ladder runs had started and before any affected reply was read.
\item The explicit instruction, the stated reference, the calculator, black-box access and the scripted baselines were each designed after earlier results were known, and each was committed before its own runs.
\item The description of near misses in the audit (the right groups named, but defined too narrowly) is post hoc.
\end{enumerate}

\section{The explicit system instruction}
\label{app:instruction}

\begin{quote}\small
Answer the user's request in this one reply; there is no one to ask and no follow-up. Infer what the request is for and carry it to completion. Where it leaves open something that would change the answer, such as a reference, a population, a tolerance or a size, choose the most defensible option, name it in one line, and continue; never stop because a choice has more than one defensible answer or because the request cannot be met exactly. If it cannot be met exactly, deliver the closest approximation and say how close it is. Give your best estimate in the requested format; an empty or unchanged answer is right only when it is your best estimate, not when the evidence is merely incomplete. Do not add warnings, disclaimers or statements about what cannot be proven.
\end{quote}

\section{Example prompt (reading, calculator, white box)}
\label{app:prompt}

\begin{quote}\small
A neural network was trained to predict death within one month in adults admitted with aneurysmal subarachnoid haemorrhage (aSAH). Its weights are loaded in your run\_python tool: W1, b1, W2, b2, \ldots{} hold them, and predict(X) gives the network's log-odds for an array of inputs in the order listed. Explain what this network actually computes, from its weights rather than from what such a model would usually show, and write a simple model that reproduces its predictions as closely as you can. There will be no follow-up messages.

Why this matters: a trained model often arrives without its data, and it may contain relationships that nobody expected. An explanation is useful only if it describes the model that was given; your simple model will be checked against the network's own predictions.

[The 20 inputs with their coding; the architecture; the requested JSON block with effects, unexpected relationships, a surrogate of linear, hinge and product terms, and the model's confidence that the network was trained on this outcome.]
\end{quote}
The prompts used in the runs, with all replies, are in the repository.

\end{document}

%% file: tables/reading.tex
\begin{tabular}{llcccc}
\toprule
Network (weights) & Planted & Default, no tool & Explicit, no tool & Calculator, white box & Calculator, black box \\
\midrule
S (353) & Threshold & 0.89--0.91; 0.41--0.89 [3/3] & -- & -- & -- \\
S (353) & Interaction & 0.56--0.81; 0.84--0.86 [3/3] & -- & 0.85--1.03; 0.88 [3/3] & -- \\
S (353) & Threshold, half strength & 0.96--1.07; 0.82--0.84 [3/3] & -- & -- & -- \\
S (353) & Interaction, half strength & 0.65--0.70; 0.75--0.79 [3/3] & -- & -- & -- \\
S (353) & None (null) & 0.89--0.90 [0/3] & -- & 0.91 [0/3] & -- \\
\addlinespace
M (1,217) & Threshold & 0.44--1.69; 0.33--0.90 [2/3] & -- & -- & -- \\
M (1,217) & Interaction & 0.51--0.66; 0.89--0.92 [3/3] & -- & -- & -- \\
M (1,217) & None (null) & $-$0.21 to 0.87 [0/3] & -- & -- & -- \\
\addlinespace
L (3,457) & Threshold & 0.52--1.43; 0.48--0.88 [3/3] & 0.40--1.41; 0.52--0.90 [2/3] & 1.02--1.03; 0.96 [3/3] & 1.03--1.07; 0.96 [3/3] \\
L (3,457) & Interaction & 0.00--2.39; 0.33--0.39 [1/3] & 0.00--0.73; 0.36--0.86 [2/3] & 0.97--1.03; 0.92 [3/3] & 1.02--1.10; 0.92 [3/3] \\
L (3,457) & None (null) & $-$0.54 to 0.88 [0/3] & 0.84--0.90 [0/3] & 0.94 [0/3] & 0.94 [0/3] \\
\addlinespace
D3 (2,817) & Threshold & -- & -- & 1.06--1.11; 0.94 [3/3] & -- \\
D3 (2,817) & Interaction & -- & -- & 0.94--1.01; 0.92 [3/3] & -- \\
D3 (2,817) & None (null) & -- & -- & 0.93--0.94 [0/3] & -- \\
\addlinespace
XL (11,009) & Threshold & -- & -- & 1.01--1.03; 0.95--0.96 [3/3] & -- \\
XL (11,009) & Interaction & -- & -- & 1.07; 0.97 [3/3] & -- \\
XL (11,009) & None (null) & -- & -- & 0.97 [0/3] & -- \\
\addlinespace
XXL (38,401) & Threshold & -- & -- & 1.00--1.02; 0.96 [3/3] & 1.02--1.04; 0.96 [3/3] \\
XXL (38,401) & Interaction & -- & -- & 0.98--1.08; 0.98 [3/3] & 0.97; 0.98 [3/3] \\
XXL (38,401) & None (null) & -- & -- & 0.98 [0/3] & 0.98 [0/3] \\
\bottomrule
\end{tabular}

%% file: tables/audit.tex
\begin{tabular}{lcccc}
\toprule
Condition & Biased: found & Biased: declined & Best Jaccard & Unbiased: false alarm \\
\midrule
Default instructions, weights in prompt & 0/3 & 3/3 & 0.00 & 0/3 \\
Default + stated additive reference & 1/3 & 2/3 & 0.00--1.00 & 2/3 \\
Explicit instructions, weights in prompt & 2/3 & 0/3 & 0.01--0.53 & 0/3 \\
Explicit + stated additive reference & 2/3 & 0/3 & 0.59--1.00 & 3/3 \\
Explicit + calculator, white box & 3/3 & 0/3 & 0.99--1.00 & 1/3 \\
Explicit + calculator, black box & 3/3 & 0/3 & 1.00 & 0/3 \\
gpt-6.1-sol + calculator, white box & 1/1 & 0/1 & 0.53 & 0/1 \\
gpt-6-sol + calculator, white box & 1/1 & 0/1 & 0.52 & 1/1 \\
gpt-6-luna + calculator, white box & 0/1 & 0/1 & 0.05 & 1/1 \\
Script, regression tree (no LLM) & 0/1 & 0/1 & 0.00 & 0/1 \\
Script, exhaustive search (no LLM) & 1/1 & 0/1 & 0.53 & 0/1 \\
\bottomrule
\end{tabular}

%% file: tables/repair.tex
\begin{tabular}{lcccccc}
\toprule
Condition & Runs & Declined & Invalid & Change in subgroup & Mean $|$change$|$ elsewhere & Bias after \\
\midrule
E8 repair, default (subgroup not given) & 3 & 2 & 0 & -- & 0.00 & $-$1.61 to $-$1.60 \\
E8 repair, explicit (subgroup not given) & 3 & 0 & 0 & -- & 0.14 & $-$1.44 \\
Subgroup given, default & 3 & 3 & 0 & 0.00 & 0.00 & $-$1.60 \\
Subgroup given, explicit + tolerance & 3 & 0 & 0 & 1.24--1.40 & 0.07--0.09 & $-$0.37 to $-$0.20 \\
+1.6 exact, default & 3 & 3 & 0 & 0.00 & 0.00 & $-$1.60 \\
+1.6 exact, explicit & 3 & 0 & 0 & 0.80 & 0.80 & $-$0.81 \\
+1.6 exact, explicit + tolerance & 3 & 0 & 2 & 0.00--2.00 & 0.00--0.70 & $-$1.60 to 0.39 \\
+1.6 exact, explicit + tolerance + calculator & 3 & 0 & 0 & 1.48--1.49 & 0.20--0.21 & $-$0.12 to $-$0.11 \\
\midrule
Script: gradient fine-tuning (207/209 parameters) & 1 & 0 & 0 & 1.20 & 0.10 & $-$0.41 \\
Script: scripted gate + least squares (29/209 parameters) & 1 & 0 & 0 & 1.51 & 0.13 & $-$0.10 \\
\bottomrule
\end{tabular}

%% file: tables/models.tex
\begin{tabular}{lccccc}
\toprule
Model & \multicolumn{2}{c}{No tool, 105-weight networks (E13)} & \multicolumn{3}{c}{Calculator, white box (E19)} \\
\cmidrule(lr){2-3}\cmidrule(lr){4-6}
 & Found & $R^2$ & Interaction ratio (L, XXL) & Null $R^2$ (L, XXL) & Audit (Jaccard; false alarm) \\
\midrule
gpt-6-astra & 6/6 & 0.94--0.97 & 0.85--1.08 & 0.91--0.98 & 0.99--1.00; 1/3 \\
gpt-6.1-sol & 3/3 & 0.96--0.97 & 0.89--0.97 & 0.94--0.98 & 0.53; 0/1 \\
gpt-6-sol & 2/3 & 0.30--0.91 & 0.95--0.99 & 0.92--0.98 & 0.52; 1/1 \\
gpt-6-luna & 2/3 & 0.49--0.84 & 0.99--1.13 & 0.93--0.97 & 0.05; 1/1 \\
gpt-5.6-sol & 3/3 & 0.33--0.93 & -- & -- & -- \\
\bottomrule
\end{tabular}

%% file: tables/prespec.tex
\begin{tabular}{llll}
\toprule
Experiment & Plan commit & Committed (UTC+8) & First run started \\
\midrule
Planted relationships in 105-weight networks & \texttt{3a6bd3c} & 2026-11-04 20:40 & 2026-11-04 20:42 \\
Size ladder, half strength and network groups; biased-label audit and repair & \texttt{e259471} & 2026-11-05 00:35 & 2026-11-05 00:35 \\
Function-preserving rewrites & \texttt{449c348} & 2026-11-05 00:42 & 2026-11-05 03:21 \\
Addendum: claims at zero locations and exact scan & \texttt{09a62cd} & 2026-11-05 00:51 & script, after the commit \\
Other models without tools (105-weight prompts) & \texttt{5bc52e3} & 2026-11-05 00:59 & 2026-11-05 00:59 \\
Stated reference for audit and repair & \texttt{1425aa0} & 2026-11-05 01:03 & 2026-11-05 01:03 \\
Explicit system instruction & \texttt{8b3a434} & 2026-11-05 01:14 & 2026-11-05 01:23 \\
Calculation-only tool, white box & \texttt{796eb7e} & 2026-11-05 02:05 & 2026-11-05 02:13 \\
Calculation-only tool, black box & \texttt{e49843a} & 2026-11-05 03:55 & 2026-11-05 04:00 \\
Other models with the tool; first repeats & \texttt{0004706} & 2026-11-05 05:55 & 2026-11-05 05:55 \\
Scripted audit without an LLM & \texttt{b759c7b} & 2026-11-05 08:39 & script, after the commit \\
Further repeats; scripted repairs without an LLM & \texttt{ab5e8f7} & 2026-11-05 08:52 & 2026-11-05 08:52 \\
Repeats so that every cell has three runs & \texttt{650e180} & 2026-11-06 13:32 & 2026-11-06 13:33 \\
\bottomrule
\end{tabular}